\documentclass[conference,10pt]{IEEEtran}
\usepackage[utf8]{inputenc}
\usepackage[numbers,sort&compress,square]{natbib}
\usepackage[backref=page]{hyperref}
\hypersetup{
colorlinks   = true,
citecolor    = blue,
urlcolor    =  blue
}
\usepackage{url}
\usepackage{graphicx}
\usepackage{amsmath,amssymb,amsfonts}
\usepackage{algorithmic}
\usepackage{textcomp}
\usepackage{xcolor}
\usepackage{multicol}
\usepackage{multirow}
\usepackage{hhline}
\usepackage[nomessages]{fp}
\usepackage{siunitx}
\newcommand{\numT}[1]{$\num[group-separator={,}]{#1}$}
\usepackage{pgf-pie} 
\usepackage{graphicx}
\usepackage{pgfplots}
\newcommand{\footURL}[1]{\footnote{\url{#1}}}
\usepackage{tabularx}
\usepackage{mathtools}

\def\BibTeX{{\rm B\kern-.05em{\sc i\kern-.025em b}\kern-.08em
    T\kern-.1667em\lower.7ex\hbox{E}\kern-.125emX}}

\begin{document}

\title{Seeking Sinhala Sentiment: Predicting Facebook Reactions of Sinhala Posts}
\author{\IEEEauthorblockN{Vihanga Jayawickrama\IEEEauthorrefmark{1}, Gihan Weeraprameshwara\IEEEauthorrefmark{1}, Nisansa de Silva\IEEEauthorrefmark{1}, Yudhanjaya Wijeratne\IEEEauthorrefmark{2}}
\IEEEauthorblockA{\IEEEauthorrefmark{1}\textit{Department of Computer Science \& Engineering} \\
\textit{University of Moratuwa}}
\IEEEauthorblockA{\IEEEauthorrefmark{2}LIRNEasia}
\IEEEauthorblockA{vihangadewmini.17@cse.mrt.ac.lk}
}

\maketitle

\begin{abstract}
The Facebook network allows its users to record their reactions to text via a typology of emotions. This network, taken at scale, is therefore a prime data set of annotated sentiment data. This paper uses millions of such reactions, derived from a decade worth of Facebook post data centred around a Sri Lankan context, to model an \textit{eye of the beholder} approach to sentiment detection for online Sinhala textual content. Three different sentiment analysis models are built, taking into account a limited subset of reactions, all reactions, and another that derives a positive/negative \textit{star rating} value. The efficacy of these models in capturing the reactions of the observers are then computed and discussed. The analysis reveals that binary classification of reactions, for Sinhala content, is significantly more accurate than the other approaches. Furthermore, the inclusion of the \textit{like} reaction hinders the capability of accurately predicting other reactions.  %
\end{abstract}

\renewcommand\IEEEkeywordsname{Keywords}


\section{Introduction}
\label{Sec:Intro}
Understanding human emotions is an interesting, yet complex process which researchers and scientists around the world have been attempting to standardize for a long period of time. In the computational sciences, sentiment analysis has become a major research topic, especially in relation to textual content~\cite{gamage2018fast, melville2009sentiment}. 

Sentiment analysis of textual content can be approached in two ways: 
1) through the perspective of the creator 2) through the perspective of the observer. Many research projects attempt to follow the first approach, but few have followed the second. Exploring the perspective of the observer would be quite important since the emotional reaction of the \textit{author} and the \textit{reader} to the same content is not necessarily identical. Much effort is generally expended in the field of political polling, for example, where the public perception of a speech is studied to assess impact. 

To the extent of our knowledge, no attempt has been made to do such analysis in Sinhala, the subject of this study. Sinhala, similar to many other regional languages, suffers from resource poverty~\cite{wijeratne2019natural}. Previous research and resources available for NLP in Sinhala are limited and isolated~\cite{de2019survey}. This is therefore an experimental attempt in bridging this knowledge gap. The objective is to predict the sentimental reaction of Facebook users to textual content posted on Facebook. This study uses a raw corpus of Sinhala Facebook posts scraped through Crowdtangle\footURL{https://www.crowdtangle.com/} by~\citet{wijeratne2020sinhala}, and analyzes the user reactions therein as a sentiment annotation that reflects the emotional reaction of a reader to the said post\cite{graziani2019jointly}. 

Overall, three models were created and tested. For the first model, a reaction vector was created for each post with the normalized reaction counts belonging to \textit{Love}, \textit{Wow}, \textit{Haha}, \textit{Sad}, and \textit{Angry} categories. \textit{Like} and \textit{Thankful}, which are outliers at positive and negative ends of the spectrum respectively, were ignored.The results showed that the procedure could predict reaction vectors with F1 scores ranging between $0.13$ and $0.52$. The second model was highly similar to the first, the only difference being the inclusion of \textit{Like} and \textit{Thankful} reactions. The resultant F1 scores ranged between $0.00$ and $0.96$.
 
In the third model, the reactions were combined to create a positivity/negativity value for each post, following the procedure presented by~\citet{de2014sensing}. Here, \textit{Love} and \textit{Wow} were considered as positive, \textit{Sad} and \textit{Angry} were considered as negative, and \textit{Haha} was ignored due to its conflicting use cases. The F1 score of this star rating value ranged between $0.29$ and $0.30$. In contrast, the binary categorization of reactions as \textit{Positive} and \textit{Negative} exhibited promising results, with F1 scores in the range $0.70$ - $0.71$ for \textit{Positive} and $0.41$ - $0.42$ for \textit{Negative},

Thus, it can be concluded that such a binary categorization system captures the sentimental reaction to Facebook post more efficiently in comparison to the multi-category reaction value system, and presents a measure of reasonable accuracy in the imputation of such sentiment.

\section{Background} 
\label{Sec:Backgr}
Many of the studies on sentiment analysis are focused on purposes such as understanding the quality of reviews given for products presented in e-commerce sites~\cite{de2014sensing, singh2017predicting} or understanding the political preferences of people \cite{caetano2018using, rudkowsky2018more}. 

Among the research on review analysis, the work of~\citet{de2014sensing} is prominent since the study had taken a path to determine sentiments on an aspect level. Different aspects were extracted from the review, and for each aspect, the sentiment value was calculated. Further, the study provides a set of guidelines to determine the semantic orientation of a subject using a sentiment lexicon, all of which are important to convert sentiment into mathematical figures. The methodology is crucial for this study since it provides the basis of one of the two workflows we discuss in this study to predict reactions for Sinhala text. 

The work of~\citet{singh2017predicting} has used several textual features such as ease of reading, subjectivity, polarity, and entropy to predict the helpfulness ratio. The model intends to assist the process of assigning a helpfulness value to a review as soon as the review is posted, thus giving the spotlight to useful reviews over irrelevant reviews, 
highlighting the usefulness of understanding the reaction of the reader to different content. Studies on political preferences cover a massive area. Many governments and political parties use social media to understand the audience.

The research done by~\citet{caetano2018using} and~\citet{rudkowsky2018more} explain two different cases where sentiment analysis is utilized in politics. \citeauthor{caetano2018using} attempts to analyze twitter data and define the homophily of the twitter audience while~\citeauthor{rudkowsky2018more} demonstrates the usability of word embedding over bag-of-words by developing a negative sentiment detection model for parliament speeches. \citeauthor{caetano2018using} concludes that the homophily level increased with the multiplex connection of the audience, while \citeauthor{rudkowsky2018more} states that the negativity of the speeches of a parliament member correlates to the position he holds in the parliament. 

The potential of Facebook data for sentiment analysis has been researched previously for different purposes. The work by~\citet{pool2016distant} and~\citet{freeman2019shared} use data sets obtained from Facebook for emotion detection. The data scope covered through the work of~\citeauthor{freeman2019shared} lacks diversity since the research is solely focused on Scholarly articles. However,~\citeauthor{pool2016distant} have attempted to maintain a general data set by using a variety of sources. The motivation behind this wide range of sources was to pick the best sources to train ML models for each reaction. \citeauthor{pool2016distant} too has looked into developing models with different features such as TF-IDF, embeddings, and n-grams. This comparison provides useful guidelines for picking up certain features in data. One of the most important aspects of the work by~\citeauthor{pool2016distant} is that they have taken an extra step to test the models with external data sets such as AffectiveText~\cite{strapparava-mihalcea-2007-semeval} and Fairy Tales~\cite{alm2008affect}, to prove the validity of the developed model since those are widely used data sets in the field of sentiment analysis. This provides a common ground to compare different sentiment analysis models. 

While all papers mentioned above provide quite useful information, almost all of them are related to English, which is a resource-rich language. On the contrary, our project will be based on the Sinhala language. This poses a major challenge to our work due to the scarcity of similar work in the domain~\cite{de2019survey} and issues with the quality of the available data~\citet{caswell2021quality}.

Among the currently available research in this arena,~\citet{senevirathne2020sentiment} is the state-of-the-art Sinhala text sentiment analysis attempt to the best of our knowledge. Through this paper, \citeauthor{senevirathne2020sentiment} has introduced a study of sentiment analysis models built using different deep learning techniques as well as an annotated sentiment data set consisting of 15059 Sinhala news comments. Furthermore, earlier attempts such as~\citet{medagoda2015sentiment} provides insight into utilizing resources available for languages such as English for generating progress in sentiment analysis in Sinhala. The partially automated framework for developing a sentiment lexicon for Sinhala presented through~\citet{chathuranga2019sinhala} is a noteworthy attempt at using a Part-of-Speech (PoS) tagged corpus for sentiment analysis. The authors proposed the use of adjectives tagged as positive or negative to predict the sentiment embedded in textual content. 
%

\section{Methodology}
\label{Sec:Meth}
This study was conducted using the raw corpus developed by~\citet{wijeratne2020sinhala} which consists of 1,820,930 Facebook posts created by pages popular in Sri Lanka between 01-01-2010 and 02-02-2020.

\subsection{Pre-processing}
The corpus was pre-processed by cleaning the \textit{Message} column and normalizing reaction counts. 
The character \textit{Zero Width Joiner} was replaced with a null string and other conrtol chracters were replaced by a space. The reason being that the \textit{Zero Width Joiner} was present in the middle of Sinhala words, especially when the characters \textit{rakāransaya}, \textit{yansaya}, and \textit{rēpaya} were used.

Furthermore, URLs, email addresses, user tags (format \textit{@user}), and hashtags were removed. Since only Sinhala and English words are to be considered, any words containing characters that are neither Sinhala nor ASCII were removed. The list of stop words for Sinhala developed from this corpus by~\citet{wijeratne2020sinhala} were removed next. Numerical content was removed due to their high unlikelihood to be repeated in the same sequence order. Finally, multiple continuous white spaces in the corpus were replaced with a single white space. The final cleaned corpus consisted of \numT{526732} data rows.

\subsection{Core Reaction Set Model}
\label{sec:core}

In selecting the \textit{core} reaction set, \textit{Like} and \textit{Thankful} reactions were excluded due to \textit{Like} being an outlier on the higher end and \textit{Thankful} being an outlier on the lower end. The total count of each reaction in the corpus along with their percentages are mentioned in Table~\ref{Table:Counts}. A probable reason for the abnormal behaviour of those reactions are the duration that they have been present on Facebook. \textit{Like} was the first reaction introduced to the platform, back in 2009~\cite{kincaid2009facebook}. \textit{Love}, \textit{Wow}, \textit{Haha}, \textit{Sad}, and \textit{Angry} reactions were introduced in 2016~\cite{stinson2016facebook}; however, \textit{Like} still retained its state as the default reaction which a simple click on the react button enforces. The \textit{Thankful} reaction was introduced for a short period of time and removed~\cite{pool2016distant} thus representing an insignificant portion of the dataset.

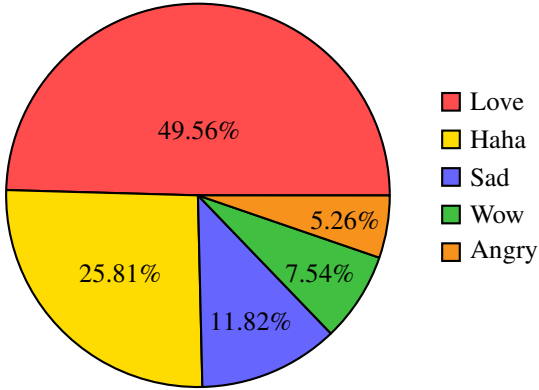
\begin{figure}[!hbt]
    \centering
    \begin{tikzpicture}
\pie[
color = {
        red!70,
        yellow!90!red,
        blue!60,
        green!50!gray,
        yellow!50!red,
    }, 
radius=2.55,
text = legend
]{\PerConLoves/Love,
    \PerConHahas/Haha,
    \PerConSads/Sad,
    \PerConWows/Wow,
    \PerConAngrys/Angry}
\end{tikzpicture}
    \caption{Core Reaction Percentages}
    \label{fig:percentages}
\end{figure}

Thus, the \textit{core} reaction set was defined considering only the \textit{Love}, \textit{Wow}, \textit{Haha}, \textit{Sad}, and \textit{Angry} reactions. The percentages of the \textit{core} reactions are also shown in Table~\ref{Table:Counts}. Furthermore, Fig.~\ref{fig:percentages} shows the \textit{core} reaction percentages as a pie chart.

\FPeval{\Likes}{clip(528060209)}
\FPeval{\Loves}{clip(12526942)}
\FPeval{\Wows}{clip(1906174)}
\FPeval{\Hahas}{clip(6524139)}
\FPeval{\Sads}{clip(2987589)}
\FPeval{\Angrys}{clip(1329552)}
\FPeval{\Thankfuls}{clip(13637)}

\FPeval{\TotalConsidered}{clip(\Loves+\Wows+\Hahas+\Sads+\Angrys)}
\FPeval{\Total}{clip(\Likes+\TotalConsidered+\Thankfuls)}

\FPeval{\PerLikes}{clip(round((100*\Likes)/\Total,3))}
\FPeval{\PerLoves}{clip(round((100*\Loves)/\Total,2))}
\FPeval{\PerWows}{clip(round((100*\Wows)/\Total,2))}
\FPeval{\PerHahas}{clip(round((100*\Hahas)/\Total,2))}
\FPeval{\PerSads}{clip(round((100*\Sads)/\Total,2))}
\FPeval{\PerAngrys}{clip(round((100*\Angrys)/\Total,2))}
\FPeval{\PerThankfuls}{clip(round((100*\Thankfuls)/\Total,3))}

\FPeval{\PerConLoves}{clip(round((100*\Loves)/\TotalConsidered,2))}
\FPeval{\PerConWows}{clip(round((100*\Wows)/\TotalConsidered,2))}
\FPeval{\PerConHahas}{clip(round((100*\Hahas)/\TotalConsidered,2))}
\FPeval{\PerConSads}{clip(round((100*\Sads)/\TotalConsidered,2))}
\FPeval{\PerConAngrys}{clip(round((100*\Angrys)/\TotalConsidered,2))}

\begin{table}[htb]
\renewcommand{\arraystretch}{1.5}
\centering
\caption{Total counts of reactions in the corpus}
\label{Table:Counts}
\begin{tabular}{|l|r|r|r|}
\hline
 \multirow{2}{*}{\textbf{Reaction}} & \multirow{2}{*}{\textbf{Count}} & \multicolumn{2}{c|}{\textbf{Percentage}} \\
 \hhline{~~--}
 & & \textbf{All} & \textbf{Core} \\
 \hline
 Like & \numT{\Likes} & \PerLikes& - \\
 Love & \numT{\Loves}& \PerLoves & \PerConLoves\\
 Wow & \numT{\Wows} & \PerWows  & \PerConWows \\
 Haha & \numT{\Hahas} & \PerHahas & \PerConHahas \\
 Sad & \numT{\Sads} & \PerSads & \PerConSads \\
 Angry & \numT{\Angrys} & \PerAngrys & \PerConAngrys \\
 Thankful & \numT{\Thankfuls} & \PerThankfuls & - \\
 \hline
\end{tabular}
\end{table}

Thus, initially, the normalization was done considering only the \textit{core} reactions. Equation~\ref{Eq:Norm1} obtains the sum of reactions ($T$) of an entry  using the counts of: \textit{Love} ($n_L$), \textit{Wow} ($n_W$), \textit{Haha} ($n_H$), \textit{Sad} ($n_S$), and \textit{Angry} ($n_A$). The Equation~\ref{Eq:Norm2} shows the normalized value $N_r$ for reaction $r$ where $n_r$ is the raw count of the reaction and $T$ is the sum obtained in Equation~\ref{Eq:Norm1}.  

\begin{equation}
    T=n_L+n_W+n_H+n_S+n_A
    \label{Eq:Norm1}
\end{equation}

\begin{equation} 
    N_r=\frac{n_r}{T}
    \label{Eq:Norm2}
\end{equation}

 The message column was then tokenized into individual words, and set operation was used to obtain the collection of unique words for each entry. A dictionary was created for each entry by assigning the normalized reaction vector of the entry to each word. The dictionaries thus created were merged vertically, taking the average value of vectors assigned to a word across the data set as the aggregate reaction vector of that word. Equation \ref{Eq:WordVec} describes this process where $V_W$ is the aggregate reaction vector for the word $W$, $R_i$ is the reaction vector of the $i$th entry ($E_i$), $n$ is the number of entries, and $\emptyset$ is the empty vector.

\begin{equation}
    V_{W} = \frac{\sum_{i=1}^{n}\begin{cases}
    R_i & \text{if } W \in E_i \\
    \emptyset & \text{otherwise}
    \end{cases}
    }
    {
    \sum_{i=1}^{n}\begin{cases}
    1 & \text{if } W \in E_i \\
    0 & \text{otherwise}
    \end{cases}
    }
\label{Eq:WordVec}
\end{equation}

The dictionary thus created was used to predict the reaction vectors of the test data set. For entries of which none of the words were found in the dictionary, the mean vector value of the train data set was assigned. Equation~\ref{Eq:MsgVec} shows the calculation of the predicted vector $V_M$ for a message where, $V_W$ is taken from the dictionary (populated as in Equation~\ref{Eq:WordVec}), and $N_M$ is the number of words in the message $M$.

\begin{equation}
    V_M = \frac{\sum_{N_M}{V_W}}{N_M}
\label{Eq:MsgVec}
\end{equation}

\subsection{Defining the Evaluation Statistics}
\label{sec:eval}

To evaluate the performance of the prediction process, a number of statistics were calculated. Equation~\ref{Eq:Ac} shows the calculation of \textit{Accuracy} $A_r$ for reaction $r$ where, $N_r$ is the expected (actual) value for the entry as calculated in Equation~\ref{Eq:Norm2} and $M_r$ is the predicted value calculated in Equation~\ref{Eq:MsgVec} as $M_r \in V_M$.  

\begin{equation}
    A_r=min \big( N_r, M_r \big)
    \label{Eq:Ac}
\end{equation}

The accuracy can be defined this way since we are solving a bin packing problem and the vector values are sum up to $1$. Standard formulas were used for calculating  \textit{Recall} ($R_r$), \textit{Precision} ($P_r$), and \textit{F1 score} ($F1_r$).

The above measures were calculated for each entry of the data set and the average of each measure was assigned as the resultant performance measure of the data set. Those values were then averaged across $5$ runs of the code.

\FPeval{\LoveAccuracyXA}{clip(round(0.3118805783,4))}
\FPeval{\WowAccuracyXA}{clip(round(0.02984379443,4))}
\FPeval{\HahaAccuracyXA}{clip(round(0.1163359895,4))}
\FPeval{\SadAccuracyXA}{clip(round(0.04967131001,4))}
\FPeval{\AngryAccuracyXA}{clip(round(0.01750127084,4))}

\FPeval{\LoveRecallXA}{clip(round(0.5862589224,4))}
\FPeval{\WowRecallXA}{clip(round(0.3110854911,4))}
\FPeval{\HahaRecallXA}{clip(round(0.4241391329,4))}
\FPeval{\SadRecallXA}{clip(round(0.2354637537,4))}
\FPeval{\AngryRecallXA}{clip(round(0.2059048856,4))}

\FPeval{\LovePrecisionXA}{clip(round(0.7837662508,4))}
\FPeval{\WowPrecisionXA}{clip(round(0.637283656,4))}
\FPeval{\HahaPrecisionXA}{clip(round(0.6279349023,4))}
\FPeval{\SadPrecisionXA}{clip(round(0.6206434824,4))}
\FPeval{\AngryPrecisionXA}{clip(round(0.5836968263,4))}

\FPeval{\LoveFXA}{clip(round(0.5164053227,4))}
\FPeval{\WowFXA}{clip(round(0.2217574193,4))}
\FPeval{\HahaFXA}{clip(round(0.3060332806,4))}
\FPeval{\SadFXA}{clip(round(0.1613184214,4))}
\FPeval{\AngryFXA}{clip(round(0.1318310537,4))}

\FPeval{\LoveAccuracyXB}{clip(round(0.3118802496,4))}
\FPeval{\WowAccuracyXB}{clip(round(0.02987373449,4))}
\FPeval{\HahaAccuracyXB}{clip(round(0.1159772707,4))}
\FPeval{\SadAccuracyXB}{clip(round(0.04965008969,4))}
\FPeval{\AngryAccuracyXB}{clip(round(0.01738724589,4))}

\FPeval{\LoveRecallXB}{clip(round(0.58474059,4))}
\FPeval{\WowRecallXB}{clip(round(0.3110224068,4))}
\FPeval{\HahaRecallXB}{clip(round(0.4242129565,4))}
\FPeval{\SadRecallXB}{clip(round(0.2360237341,4))}
\FPeval{\AngryRecallXB}{clip(round(0.204117007,4))}

\FPeval{\LovePrecisionXB}{clip(round(0.7833380992,4))}
\FPeval{\WowPrecisionXB}{clip(round(0.6374574778,4))}
\FPeval{\HahaPrecisionXB}{clip(round(0.6260624964,4))}
\FPeval{\SadPrecisionXB}{clip(round(0.6205300494,4))}
\FPeval{\AngryPrecisionXB}{clip(round(0.58339977,4))}

\FPeval{\LoveFXB}{clip(round(0.5147420945,4))}
\FPeval{\WowFXB}{clip(round(0.2216459881,4))}
\FPeval{\HahaFXB}{clip(round(0.305273546,4))}
\FPeval{\SadFXB}{clip(round(0.1615503853,4))}
\FPeval{\AngryFXB}{clip(round(0.1308299319,4))}

\FPeval{\LoveAccuracyXC}{clip(round(0.3118120159,4))}
\FPeval{\WowAccuracyXC}{clip(round(0.02980374253,4))}
\FPeval{\HahaAccuracyXC}{clip(round(0.1160425227,4))}
\FPeval{\SadAccuracyXC}{clip(round(0.04991370153,4))}
\FPeval{\AngryAccuracyXC}{clip(round(0.01736700842,4))}

\FPeval{\LoveRecallXC}{clip(round(0.5854000432,4))}
\FPeval{\WowRecallXC}{clip(round(0.3113394516,4))}
\FPeval{\HahaRecallXC}{clip(round(0.4238044234,4))}
\FPeval{\SadRecallXC}{clip(round(0.2379614239,4))}
\FPeval{\AngryRecallXC}{clip(round(0.2044779098,4))}

\FPeval{\LovePrecisionXC}{clip(round(0.7832681511,4))}
\FPeval{\WowPrecisionXC}{clip(round(0.6369907289,4))}
\FPeval{\HahaPrecisionXC}{clip(round(0.6265789643,4))}
\FPeval{\SadPrecisionXC}{clip(round(0.6175779876,4))}
\FPeval{\AngryPrecisionXC}{clip(round(0.5855846267,4))}

\FPeval{\LoveFXC}{clip(round(0.5153391631,4))}
\FPeval{\WowFXC}{clip(round(0.2217515481,4))}
\FPeval{\HahaFXC}{clip(round(0.3052284622,4))}
\FPeval{\SadFXC}{clip(round(0.16225395,4))}
\FPeval{\AngryFXC}{clip(round(0.1313548007,4))}

\FPeval{\LoveAccuracyXD}{clip(round(0.31173786934,4))}
\FPeval{\WowAccuracyXD}{clip(round(0.02982159471,4))}
\FPeval{\HahaAccuracyXD}{clip(round(0.1158448039,4))}
\FPeval{\SadAccuracyXD}{clip(round(0.049659653,4))}
\FPeval{\AngryAccuracyXD}{clip(round(0.01736227281,4))}

\FPeval{\LoveRecallXD}{clip(round(0.5854859377,4))}
\FPeval{\WowRecallXD}{clip(round(0.3110179069,4))}
\FPeval{\HahaRecallXD}{clip(round(0.4236473213,4))}
\FPeval{\SadRecallXD}{clip(round(0.2368190026,4))}
\FPeval{\AngryRecallXD}{clip(round(0.2049585142,4))}

\FPeval{\LovePrecisionXD}{clip(round(0.7829151073,4))}
\FPeval{\WowPrecisionXD}{clip(round(0.6375978847,4))}
\FPeval{\HahaPrecisionXD}{clip(round(0.6262579407,4))}
\FPeval{\SadPrecisionXD}{clip(round(0.618264614,4))}
\FPeval{\AngryPrecisionXD}{clip(round(0.5846595758,4))}

\FPeval{\LoveFXD}{clip(round(0.5151606205,4))}
\FPeval{\WowFXD}{clip(round(0.221662067,4))}
\FPeval{\HahaFXD}{clip(round(0.3049049485,4))}
\FPeval{\SadFXD}{clip(round(0.1615546311,4))}
\FPeval{\AngryFXD}{clip(round(0.1314398781,4))}

\FPeval{\LoveAccuracyXE}{clip(round(0.3120539517,4))}
\FPeval{\WowAccuracyXE}{clip(round(0.02975576251,4))}
\FPeval{\HahaAccuracyXE}{clip(round(0.1155302273,4))}
\FPeval{\SadAccuracyXE}{clip(round(0.04959642111,4))}
\FPeval{\AngryAccuracyXE}{clip(round(0.0172909434,4))}

\FPeval{\LoveRecallXE}{clip(round(0.5863042214,4))}
\FPeval{\WowRecallXE}{clip(round(0.3112857871,4))}
\FPeval{\HahaRecallXE}{clip(round(0.4236453919,4))}
\FPeval{\SadRecallXE}{clip(round(0.2366007586,4))}
\FPeval{\AngryRecallXE}{clip(round(0.2041202857,4))}

\FPeval{\LovePrecisionXE}{clip(round(0.7823548059,4))}
\FPeval{\WowPrecisionXE}{clip(round(0.6361154851,4))}
\FPeval{\HahaPrecisionXE}{clip(round(0.6248883134,4))}
\FPeval{\SadPrecisionXE}{clip(round(0.6195023377,4))}
\FPeval{\AngryPrecisionXE}{clip(round(0.5855480664,4))}

\FPeval{\LoveFXE}{clip(round(0.5155545549,4))}
\FPeval{\WowFXE}{clip(round(0.2213736234,4))}
\FPeval{\HahaFXE}{clip(round(0.304302757,4))}
\FPeval{\SadFXE}{clip(round(0.1616958892,4))}
\FPeval{\AngryFXE}{clip(round(0.1310042641,4))}



\FPeval{\LikesAccuracyYA}{clip(round(0.9169207037,4))}
\FPeval{\LoveAccuracyYA}{clip(round(0.005644154784,4))}
\FPeval{\WowAccuracyYA}{clip(round(0.0004911230282,4))}
\FPeval{\HahaAccuracyYA}{clip(round(0.004223909216,4))}
\FPeval{\SadAccuracyYA}{clip(round(0.001520312931,4))}
\FPeval{\AngryAccuracyYA}{clip(round(0.0005889384874,4))}
\FPeval{\ThankfulAccuracyYA}{clip(round(0.00000002967310208,4))}

\FPeval{\LikesRecallYA}{clip(round(0.9651310753,4))}
\FPeval{\LoveRecallYA}{clip(round(0.2509692837,4))}
\FPeval{\WowRecallYA}{clip(round(0.1487344468,4))}
\FPeval{\HahaRecallYA}{clip(round(0.1646345223,4))}
\FPeval{\SadRecallYA}{clip(round(0.1013322171,4))}
\FPeval{\AngryRecallYA}{clip(round(0.08804807051,4))}
\FPeval{\ThankfulRecallYA}{clip(round(0.0007386222413,4))}

\FPeval{\LikesPrecisionYA}{clip(round(0.9690724132,4))}
\FPeval{\LovePrecisionYA}{clip(round(0.6220837698,4))}
\FPeval{\WowPrecisionYA}{clip(round(0.4550043322,4))}
\FPeval{\HahaPrecisionYA}{clip(round(0.6044408252,4))}
\FPeval{\SadPrecisionYA}{clip(round(0.5828895539,4))}
\FPeval{\AngryPrecisionYA}{clip(round(0.5193011063,4))}
\FPeval{\ThankfulPrecisionYA}{clip(round(0.04398909321,4))}

\FPeval{\LikesFYA}{clip(round(0.9625903236,4))}
\FPeval{\LoveFYA}{clip(round(0.1769485655,4))}
\FPeval{\WowFYA}{clip(round(0.08181898095,4))}
\FPeval{\HahaFYA}{clip(round(0.1067527348,4))}
\FPeval{\SadFYA}{clip(round(0.06383285326,4))}
\FPeval{\AngryFYA}{clip(round(0.04949641528,4))}
\FPeval{\ThankfulFYA}{clip(round(0.00003683418613,4))}

\FPeval{\LikesAccuracyYB}{clip(round(0.916960193,4))}
\FPeval{\LoveAccuracyYB}{clip(round(0.005639997689,4))}
\FPeval{\WowAccuracyYB}{clip(round(0.0004900495542,4))}
\FPeval{\HahaAccuracyYB}{clip(round(0.004226092181,4))}
\FPeval{\SadAccuracyYB}{clip(round(0.001523439297,4))}
\FPeval{\AngryAccuracyYB}{clip(round(0.0005922923778,4))}
\FPeval{\ThankfulAccuracyYB}{clip(round(0.0000000165385933,4))}

\FPeval{\LikesRecallYB}{clip(round(0.9651610023,4))}
\FPeval{\LoveRecallYB}{clip(round(0.2513066234,4))}
\FPeval{\WowRecallYB}{clip(round(0.1486411138,4))}
\FPeval{\HahaRecallYB}{clip(round(0.1638620443,4))}
\FPeval{\SadRecallYB}{clip(round(0.1008646379,4))}
\FPeval{\AngryRecallYB}{clip(round(0.08821830528,4))}
\FPeval{\ThankfulRecallYB}{clip(round(0.0006974832014,4))}

\FPeval{\LikesPrecisionYB}{clip(round(0.9690551057,4))}
\FPeval{\LovePrecisionYB}{clip(round(0.6224912515,4))}
\FPeval{\WowPrecisionYB}{clip(round(0.4556886511,4))}
\FPeval{\HahaPrecisionYB}{clip(round(0.6042562588,4))}
\FPeval{\SadPrecisionYB}{clip(round(0.5840214799,4))}
\FPeval{\AngryPrecisionYB}{clip(round(0.5162401686,4))}
\FPeval{\ThankfulPrecisionYB}{clip(round(0.03761278303,4))}

\FPeval{\LikesFYB}{clip(round(0.9626183391,4))}
\FPeval{\LoveFYB}{clip(round(0.177380626,4))}
\FPeval{\WowFYB}{clip(round(0.08180976134,4))}
\FPeval{\HahaFYB}{clip(round(0.1063510121,4))}
\FPeval{\SadFYB}{clip(round(0.06356872304,4))}
\FPeval{\AngryFYB}{clip(round(0.04943573989,4))}
\FPeval{\ThankfulFYB}{clip(round(0.00003575021504,4))}

\FPeval{\LikesAccuracyYC}{clip(round(0.9167391822,4))}
\FPeval{\LoveAccuracyYC}{clip(round(0.005635261447,4))}
\FPeval{\WowAccuracyYC}{clip(round(0.0004906741018,4))}
\FPeval{\HahaAccuracyYC}{clip(round(0.004237710133,4))}
\FPeval{\SadAccuracyYC}{clip(round(0.00152975478,4))}
\FPeval{\AngryAccuracyYC}{clip(round(0.0005949014289,4))}
\FPeval{\ThankfulAccuracyYC}{clip(round(0.00000002124900305,4))}

\FPeval{\LikesRecallYC}{clip(round(0.9649457099,4))}
\FPeval{\LoveRecallYC}{clip(round(0.2514625766,4))}
\FPeval{\WowRecallYC}{clip(round(0.1489787471,4))}
\FPeval{\HahaRecallYC}{clip(round(0.164740939,4))}
\FPeval{\SadRecallYC}{clip(round(0.1011837064,4))}
\FPeval{\AngryRecallYC}{clip(round(0.08893031317,4))}
\FPeval{\ThankfulRecallYC}{clip(round(0.0006879037582,4))}

\FPeval{\LikesPrecisionYC}{clip(round(0.9690992617,4))}
\FPeval{\LovePrecisionYC}{clip(round(0.6207757502,4))}
\FPeval{\WowPrecisionYC}{clip(round(0.4527471217,4))}
\FPeval{\HahaPrecisionYC}{clip(round(0.6037438883,4))}
\FPeval{\SadPrecisionYC}{clip(round(0.5824629539,4))}
\FPeval{\AngryPrecisionYC}{clip(round(0.5142373213,4))}
\FPeval{\ThankfulPrecisionYC}{clip(round(0.02969271064,4))}

\FPeval{\LikesFYC}{clip(round(0.9625167356,4))}
\FPeval{\LoveFYC}{clip(round(0.1770198508,4))}
\FPeval{\WowFYC}{clip(round(0.08161055715,4))}
\FPeval{\HahaFYC}{clip(round(0.1067189137,4))}
\FPeval{\SadFYC}{clip(round(0.06362762872,4))}
\FPeval{\AngryFYC}{clip(round(0.04965234698,4))}
\FPeval{\ThankfulFYC}{clip(round(0.0000267902339,4))}

\FPeval{\LikesAccuracyYD}{clip(round(0.9166737769,4))}
\FPeval{\LoveAccuracyYD}{clip(round(0.005643755856,4))}
\FPeval{\WowAccuracyYD}{clip(round(0.0004882786486,4))}
\FPeval{\HahaAccuracyYD}{clip(round(0.004231934489,4))}
\FPeval{\SadAccuracyYD}{clip(round(0.001531572838,4))}
\FPeval{\AngryAccuracyYD}{clip(round(0.000589795154,4))}
\FPeval{\ThankfulAccuracyYD}{clip(round(0.00000002133003724,4))}

\FPeval{\LikesRecallYD}{clip(round(0.9649551987,4))}
\FPeval{\LoveRecallYD}{clip(round(0.2512641988,4))}
\FPeval{\WowRecallYD}{clip(round(0.1484309262,4))}
\FPeval{\HahaRecallYD}{clip(round(0.1643456612,4))}
\FPeval{\SadRecallYD}{clip(round(0.1014124643,4))}
\FPeval{\AngryRecallYD}{clip(round(0.08852484896,4))}
\FPeval{\ThankfulRecallYD}{clip(round(0.0007183385412,4))}

\FPeval{\LikesPrecisionYD}{clip(round(0.9690123119,4))}
\FPeval{\LovePrecisionYD}{clip(round(0.6216389209,4))}
\FPeval{\WowPrecisionYD}{clip(round(0.453882682,4))}
\FPeval{\HahaPrecisionYD}{clip(round(0.6044909011,4))}
\FPeval{\SadPrecisionYD}{clip(round(0.5816430521,4))}
\FPeval{\AngryPrecisionYD}{clip(round(0.5154999787,4))}
\FPeval{\ThankfulPrecisionYD}{clip(round(0.03418417666,4))}

\FPeval{\LikesFYD}{clip(round(0.9624715206,4))}
\FPeval{\LoveFYD}{clip(round(0.1770868224,4))}
\FPeval{\WowFYD}{clip(round(0.08143406276,4))}
\FPeval{\HahaFYD}{clip(round(0.1065264135,4))}
\FPeval{\SadFYD}{clip(round(0.06365503382,4))}
\FPeval{\AngryFYD}{clip(round(0.04948473456,4))}
\FPeval{\ThankfulFYD}{clip(round(0.00003237669253,4))}

\FPeval{\LikesAccuracyYE}{clip(round(0.9167081862,4))}
\FPeval{\LoveAccuracyYE}{clip(round(0.005629186094,4))}
\FPeval{\WowAccuracyYE}{clip(round(0.0004885296612,4))}
\FPeval{\HahaAccuracyYE}{clip(round(0.004215100947,4))}
\FPeval{\SadAccuracyYE}{clip(round(0.001523737447,4))}
\FPeval{\AngryAccuracyYE}{clip(round(0.0005883503268,4))}
\FPeval{\ThankfulAccuracyYE}{clip(round(0.00000001967948116,4))}

\FPeval{\LikesRecallYE}{clip(round(0.9650128385,4))}
\FPeval{\LoveRecallYE}{clip(round(0.251305879,4))}
\FPeval{\WowRecallYE}{clip(round(0.1490511653,4))}
\FPeval{\HahaRecallYE}{clip(round(0.164258149,4))}
\FPeval{\SadRecallYE}{clip(round(0.1013652716,4))}
\FPeval{\AngryRecallYE}{clip(round(0.08856069106,4))}
\FPeval{\ThankfulRecallYE}{clip(round(0.0007196079816,4))}

\FPeval{\LikesPrecisionYE}{clip(round(0.968965963,4))}
\FPeval{\LovePrecisionYE}{clip(round(0.620640087,4))}
\FPeval{\WowPrecisionYE}{clip(round(0.4518955655,4))}
\FPeval{\HahaPrecisionYE}{clip(round(0.6034428975,4))}
\FPeval{\SadPrecisionYE}{clip(round(0.5814678793,4))}
\FPeval{\AngryPrecisionYE}{clip(round(0.5141906283,4))}
\FPeval{\ThankfulPrecisionYE}{clip(round(0.0329879043,4))}

\FPeval{\LikesFYE}{clip(round(0.9624948267,4))}
\FPeval{\LoveFYE}{clip(round(0.1768238888,4))}
\FPeval{\WowFYE}{clip(round(0.08147154956,4))}
\FPeval{\HahaFYE}{clip(round(0.1062553315,4))}
\FPeval{\SadFYE}{clip(round(0.0636026422,4))}
\FPeval{\AngryFYE}{clip(round(0.04938732691,4))}
\FPeval{\ThankfulFYE}{clip(round(0.00003184971069,4))}

\begin{table*}[htb]
\renewcommand{\arraystretch}{1.5}
\centering
\caption{Performance Measures of Vector Predictions}
\label{Table:VecPredMeasure}

\begin{tabularx}{\textwidth}{
|>{\centering\arraybackslash}X
|>{\centering\arraybackslash}X
|>{\centering\arraybackslash}X
|>{\centering\arraybackslash}X
|>{\centering\arraybackslash}X
|>{\centering\arraybackslash}X
||>{\centering\arraybackslash}X
|>{\centering\arraybackslash}X
|>{\centering\arraybackslash}X
|>{\centering\arraybackslash}X
|}
\hline
 \multirow{2}{*}{\textbf{Train (\%)}} &
 \multirow{2}{*}{\textbf{Reaction}} &
 \multicolumn{4}{c||}{\textbf{Core Reaction Set Model}} & \multicolumn{4}{c|}{\textbf{All Reaction Set Model}} \\
 \hhline{~~--------}
 & & \textbf{Accuracy} & \textbf{Recall} & \textbf{Precision} & \textbf{F1 Score} & \textbf{Accuracy} & \textbf{Recall} & \textbf{Precision} & \textbf{F1 Score}\\
 \hline
 \multirow{7}{*}{95} & Like & - & - & - & - & \LikesAccuracyYA & \LikesRecallYA & \LikesPrecisionYA & \LikesFYA\\
 & Love & \LoveAccuracyXA & \LoveRecallXA & \LovePrecisionXA & \textbf{\LoveFXA} & \LoveAccuracyYA & 0.2510 & \LovePrecisionYA & \LoveFYA\\
 & Wow & \WowAccuracyXA & \WowRecallXA & \WowPrecisionXA & \textbf{\WowFXA} & \WowAccuracyYA & \WowRecallYA & 0.4550 & \WowFYA\\
 & Haha & \textbf{\HahaAccuracyXA} & \HahaRecallXA & \HahaPrecisionXA & \textbf{0.3060} & \HahaAccuracyYA & \HahaRecallYA & \HahaPrecisionYA & \HahaFYA\\
 & Sad & \SadAccuracyXA & \SadRecallXA & \SadPrecisionXA & \SadFXA & \SadAccuracyYA & \SadRecallYA & \SadPrecisionYA & \SadFYA\\
 & Angry & \textbf{\AngryAccuracyXA} & \AngryRecallXA & \AngryPrecisionXA & \textbf{\AngryFXA} & \AngryAccuracyYA & 0.0880 & \AngryPrecisionYA & \AngryFYA\\
 & Thankful & - & - & - & - & 0.0000 & \ThankfulRecallYA & 0.0440 & 0.0000 \\
 \hline
  \multirow{7}{*}{90} & Like & - & - & - & - & 0.9170 & \LikesRecallYB & \LikesPrecisionYB & \LikesFYB\\
 & Love & \LoveAccuracyXB & \LoveRecallXB & \LovePrecisionXB & \LoveFXB & \LoveAccuracyYB & \LoveRecallYB & \LovePrecisionYB & \LoveFYB\\
 & Wow & \textbf{\WowAccuracyXB} & 0.3110 & \WowPrecisionXB & \WowFXB & \WowAccuracyYB & \WowRecallYB & \WowPrecisionYB & \WowFYB\\
 & Haha & 0.1160 & \HahaRecallXB & \HahaPrecisionXB & \HahaFXB & \HahaAccuracyYB & \HahaRecallYB & \HahaPrecisionYB & \HahaFYB\\
 & Sad & \SadAccuracyXB & 0.2360 & \SadPrecisionXB & \SadFXB & \SadAccuracyYB & \SadRecallYB & 0.5840 & \SadFYB\\
 & Angry & \AngryAccuracyXB & \AngryRecallXB & \AngryPrecisionXB & \AngryFXB & \AngryAccuracyYB & \AngryRecallYB & \AngryPrecisionYB & \AngryFYB\\
 & Thankful & - & - & - & - & 0.0000 & \ThankfulRecallYB & \ThankfulPrecisionYB & 0.0000\\
 \hline
  \multirow{7}{*}{80} & Like & - & - & - & - & \LikesAccuracyYC & \LikesRecallYC & \LikesPrecisionYC & \LikesFYC\\
 & Love & \LoveAccuracyXC & \LoveRecallXC & \LovePrecisionXC & \LoveFXC & \LoveAccuracyYC & \LoveRecallYC & \LovePrecisionYC & 0.1770\\
 & Wow & \WowAccuracyXC & \WowRecallXC & 0.6370 & \textbf{\WowFXC} & \WowAccuracyYC & 0.1490 & \WowPrecisionYC & \WowFYC\\
 & Haha & 0.1160 & \HahaRecallXC & \HahaPrecisionXC & \HahaFXC & \HahaAccuracyYC & \HahaRecallYC & \HahaPrecisionYC & \HahaFYC\\
 & Sad & \textbf{\SadAccuracyXC} & 0.2380 & \SadPrecisionXC & \textbf{\SadFXC} & \SadAccuracyYC & \SadRecallYC & \SadPrecisionYC & \SadFYC\\
 & Angry & \AngryAccuracyXC & \AngryRecallXC & \AngryPrecisionXC & \AngryFXC & \AngryAccuracyYC & \AngryRecallYC & \AngryPrecisionYC & \AngryFYC\\
 & Thankful & - & - & - & - & 0.0000 & \ThankfulRecallYC & \ThankfulPrecisionYC & 0.0000\\
 \hline
  \multirow{7}{*}{70} & Like & - & - & - & - & \LikesAccuracyYD & 0.9650 & 0.9690 & \LikesFYD\\
 & Love & \LoveAccuracyXD & \LoveRecallXD & \LovePrecisionXD & \LoveFXD & \LoveAccuracyYD & \LoveRecallYD & \LovePrecisionYD & \LoveFYD\\
 & Wow & \WowAccuracyXD & 0.3110 & \WowPrecisionXD & \WowFXD & \WowAccuracyYD & \WowRecallYD & \WowPrecisionYD & \WowFYD\\
 & Haha & \HahaAccuracyXD & \HahaRecallXD & \HahaPrecisionXD & \HahaFXD & \HahaAccuracyYD & \HahaRecallYD & \HahaPrecisionYD & \HahaFYD\\
 & Sad & \SadAccuracyXD & \SadRecallXD & \SadPrecisionXD & \SadFXD & \SadAccuracyYD & \SadRecallYD & \SadPrecisionYD & \SadFYD\\
 & Angry & \AngryAccuracyXD & 0.2050 & \AngryPrecisionXD & \AngryFXD & \AngryAccuracyYD & \AngryRecallYD & \AngryPrecisionYD & \AngryFYD\\
 & Thankful & - & - & - & - & 0.0000 & \ThankfulRecallYD & \ThankfulPrecisionYD & 0.0000 \\
 \hline
  \multirow{6}{*}{50} & Like & - & - & - & - & \LikesAccuracyYE & 0.9650 & 0.9690 & \LikesFYE\\
 & Love & \textbf{\LoveAccuracyXE} & \LoveRecallXE & \LovePrecisionXE & \LoveFXE & \LoveAccuracyYE & \LoveRecallYE & \LovePrecisionYE & \LoveFYE\\
 & Wow & \WowAccuracyXE & \WowRecallXE & \WowPrecisionXE & \WowFXE & \WowAccuracyYE & \WowRecallYE & \WowPrecisionYE & \WowFYE\\
 & Haha & \HahaAccuracyXE & \HahaRecallXE & \HahaPrecisionXE & \HahaFXE & \HahaAccuracyYE & \HahaRecallYE & \HahaPrecisionYE & \HahaFYE\\
 & Sad & \SadAccuracyXE & \SadRecallXE & \SadPrecisionXE & \SadFXE & \SadAccuracyYE & \SadRecallYE & \SadPrecisionYE & \SadFYE\\
 & Angry & \AngryAccuracyXE & \AngryRecallXE & \AngryPrecisionXE & 0.1310 & \AngryAccuracyYE & \AngryRecallYE & \AngryPrecisionYE & \AngryFYE\\
 & Thankful & - & - & - & - & 0.0000 & \ThankfulRecallYE & 0.0330 & 0.0000\\
 \hline
\end{tabularx}
\end{table*}

\subsection{All Reaction Set Model}
\label{sec:all}

The all reaction set model was developed following the same procedure of the core reaction set model. In addition to the reactions included in the core reaction set, 
\textit{Like} ($n_{Li}$) and \textit{Thankful} ($n_T$), were considered during this step. Equation~\ref{Eq:Norm1A} depicts how the sum of reactions is obtained while the normalized value $N_r^{*}$ for each reaction could be obtained as mentioned in Equation~\ref{Eq:Norm2A}. $T^{*}$ refers to the sum of reactions obtained through Equation~\ref{Eq:Norm1A}.

\begin{equation}
    T^*=n_{Li}+n_L+n_W+n_H+n_S+n_A+n_T
    \label{Eq:Norm1A}
\end{equation}

\begin{equation}
    N_r^{*}=\frac{n_r}{T^*}
    \label{Eq:Norm2A}
\end{equation}

The sentiment vector for each entry was then generated following the same procedure as in~\ref{sec:core}. The evaluation was done as mentioned in~\ref{sec:eval}.

\subsection{Star Rating Model}
\label{sec:star}

The next step was inspired by the procedure proposed by \citet{de2014sensing}. They propose using the star rating to generate sentiment vectors. Since the star rating takes a value between 1 and 5 where 3 is considered neutral, and values more than 3 and less than 3 are considered as positive and negative respectively by them. To adjust Facebook reactions to this scale, we classified the positivity of reactions as presented in table \ref{Table:PosNegReacts}. The positivity of the \textit{Haha} reaction is considered to be uncertain due to its conflicting use cases: the reaction is often used both genuinely and sarcastically on the platform~\cite{kuo2018facebook}. 

\begin{table}[htb]
\renewcommand{\arraystretch}{1.5}
\centering
\caption{Positivity and Negativity of Facebook Reactions}
\label{Table:PosNegReacts}
\begin{tabular}{|c|c|}
\hline
\textbf{Reaction} & \textbf{Positivity/Negativity}\\
\hline
Love & Positive\\
Wow & Positive\\
Haha & Uncertain\\
Sad & Negative\\
Angry & Negative\\
\hline
\end{tabular}
\end{table}

Therefore, the experiment was carried out considering only the \textit{Love}, \textit{Wow}, \textit{Sad}, and \textit{Angry} reactions. The normalization process described in Section~\ref{sec:core} for the \textit{Core Reaction Set Model} was updated by modifying Equations~\ref{Eq:Norm1} and~\ref{Eq:Norm2} as was done in Section~\ref{sec:all}. Figure \ref{fig:AmznPie} presents the distribution of selected reactions in the corpus.

\FPeval{\TotalConsideredAmzn}{clip(\Loves+\Wows+\Sads+\Angrys)}

\FPeval{\PerConLovesAmzn}{clip(round((100*\Loves)/\TotalConsideredAmzn,2))}
\FPeval{\PerConWowsAmzn}{clip(round((100*\Wows)/\TotalConsideredAmzn,2))}
\FPeval{\PerConSadsAmzn}{clip(round((100*\Sads)/\TotalConsideredAmzn,2))}
\FPeval{\PerConAngrysAmzn}{clip(round((100*\Angrys)/\TotalConsideredAmzn,2))}

\newcommand{\StarRadius}{2.85}

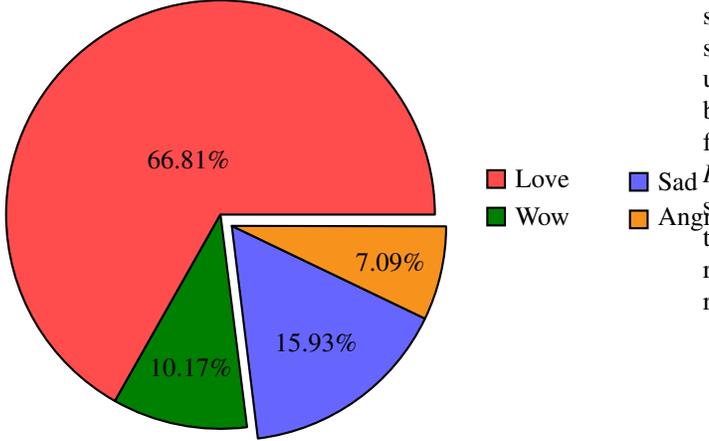
\begin{figure}
    \centering
    \begin{tikzpicture}
\pie[
color = {
        red!70,
        green!50!black,
    }, 
radius=\StarRadius,
text = legend
]{\PerConLovesAmzn/Love,
    \PerConWowsAmzn/Wow
    }
    \pie[
    pos={0.15,-0.15},
    color = {
       blue!60,
        yellow!50!red,
    },
    rotate = 277,
    radius=\StarRadius,
    text = legend,
    sum = 100
    ]{
    \PerConSadsAmzn/Sad,
    \PerConAngrysAmzn/Angry
    }
\end{tikzpicture}
    \caption{Reactions Considered for the Star Rating Model}
    \label{fig:AmznPie}
\end{figure}

The positive sentiment value ($E_{(P,i)}$) for entry $i$ was calculated by summing the \textit{Normalized Love} ($N_L^{'}$) and \textit{Normalized Wow} ($N_W^{'}$) values while the negative sentiment ($E_{(N,i)}$) was calculated by summing the \textit{Normalized Sad} ($N_S^{'}$) and \textit{Normalized Angry} ($N_A^{'}$) values. Using $E_{(P,i)}$ and $E_{(N,i)}$, the aggregated sentiment for entry $i$ was calculated as shown in Equation \ref{Eq:AggrSent}.

\begin{equation}
E_{i}=E_{(P,i)}-E_{(N,i)}
\label{Eq:AggrSent}
\end{equation}

The \textit{Star Rating Value} ($S_i^{'}$) for entry $i$ which is calculated over the entire data set was computed as shown in Equation \ref{Eq:Star} where $I$ is the set of entries in the data set.

\begin{equation}
S_i=4 * \Bigg(\frac{E_{i}-\underset{E_{j} \in I}{\mathrm{min}}(E_{j})}{\underset{E_{j} \in I}{\mathrm{max}}(E_{j})-\underset{E_{j} \in I}{\mathrm{min}}(E_{j})}\Bigg) + 1
\label{Eq:Star}
\end{equation}

The sentiment vector ($V_i$) for entry $i$ is defined in Equation~\ref{Eq:SenVec} where $E_{(P,i)}$,  $E_{(N,i)}$, $S_i^{'}$, and $S_i$ were calculated as mentioned before.

\begin{equation}
    V_i = \big[ E_{(P,i)}, E_{(N,i)}, S_i^{'}, S_i \big]
\label{Eq:SenVec}
\end{equation}

Once the vectors were computed, the processing of test and train sets, building of the dictionary, and evaluating the model was conducted akin to that in Section~\ref{sec:eval} and Section~\ref{sec:core}. The performance measures of the model were calculated using Gaussian distances.

\FPeval{\PositiveAccuracyAA}{clip(round(0.5405772358,4))}
\FPeval{\NegativeAccuracyAA}{clip(round(0.2062481067,4))}
\FPeval{\StarRatingAccuracyAA}{clip(round(0.6930276727,4))}

\FPeval{\PositiveRecallAA}{clip(round(0.7495664434,4))}
\FPeval{\NegativeRecallAA}{clip(round(0.4775483796,4))}
\FPeval{\StarRatingRecallAA}{clip(round(0.6911743266,4))}

\FPeval{\PositivePrecisionAA}{clip(round(0.8600622251,4))}
\FPeval{\NegativePrecisionAA}{clip(round(0.8067367166,4))}
\FPeval{\StarRatingPrecisionAA}{clip(round(0.2259373599,4))}

\FPeval{\PositiveFAA}{clip(round(0.706758264,4))}
\FPeval{\NegativeFAA}{clip(round(0.4206628122,4))}
\FPeval{\StarRatingFAA}{clip(round(0.292125644,4))}

\FPeval{\PositiveAccuracyAB}{clip(round(0.5420339949,4))}
\FPeval{\NegativeAccuracyAB}{clip(round(0.2052197199,4))}
\FPeval{\StarRatingAccuracyAB}{clip(round(0.6930820776,4))}

\FPeval{\PositiveRecallAB}{clip(round(0.7524158488,4))}
\FPeval{\NegativeRecallAB}{clip(round(0.4753474077,4))}
\FPeval{\StarRatingRecallAB}{clip(round(0.6913129814,4))}

\FPeval{\PositivePrecisionAB}{clip(round(0.8589015733,4))}
\FPeval{\NegativePrecisionAB}{clip(round(0.8069303146,4))}
\FPeval{\StarRatingPrecisionAB}{clip(round(0.2266580188,4))}

\FPeval{\PositiveFAB}{clip(round(0.7088493434,4))}
\FPeval{\NegativeFAB}{clip(round(0.4191851226,4))}
\FPeval{\StarRatingFAB}{clip(round(0.2945172625,4))}

\FPeval{\PositiveAccuracyAC}{clip(round(0.5415558312,4))}
\FPeval{\NegativeAccuracyAC}{clip(round(0.2037756577,4))}
\FPeval{\StarRatingAccuracyAC}{clip(round(0.6917311535,4))}

\FPeval{\PositiveRecallAC}{clip(round(0.7527207225,4))}
\FPeval{\NegativeRecallAC}{clip(round(0.4717749246,4))}
\FPeval{\StarRatingRecallAC}{clip(round(0.6895720454,4))}

\FPeval{\PositivePrecisionAC}{clip(round(0.8570604692,4))}
\FPeval{\NegativePrecisionAC}{clip(round(0.8076980475,4))}
\FPeval{\StarRatingPrecisionAC}{clip(round(0.2235990998,4))}

\FPeval{\PositiveFAC}{clip(round(0.7075051628,4))}
\FPeval{\NegativeFAC}{clip(round(0.4158581013,4))}
\FPeval{\StarRatingFAC}{clip(round(0.2912459877,4))}

\FPeval{\PositiveAccuracyAD}{clip(round(0.5409844208,4))}
\FPeval{\NegativeAccuracyAD}{clip(round(0.204631198,4))}
\FPeval{\StarRatingAccuracyAD}{clip(round(0.6924772559,4))}

\FPeval{\PositiveRecallAD}{clip(round(0.7502600016,4))}
\FPeval{\NegativeRecallAD}{clip(round(0.4751330331,4))}
\FPeval{\StarRatingRecallAD}{clip(round(0.6904561217,4))}

\FPeval{\PositivePrecisionAD}{clip(round(0.8587751853,4))}
\FPeval{\NegativePrecisionAD}{clip(round(0.8051024746,4))}
\FPeval{\StarRatingPrecisionAD}{clip(round(0.2280210256,4))}

\FPeval{\PositiveFAD}{clip(round(0.7064676687,4))}
\FPeval{\NegativeFAD}{clip(round(0.4175748516,4))}
\FPeval{\StarRatingFAD}{clip(round(0.2974866944,4))}

\FPeval{\PositiveAccuracyAE}{clip(round(0.5403207491,4))}
\FPeval{\NegativeAccuracyAE}{clip(round(0.2039560873,4))}
\FPeval{\StarRatingAccuracyAE}{clip(round(0.6914822449,4))}

\FPeval{\PositiveRecallAE}{clip(round(0.751407986,4))}
\FPeval{\NegativeRecallAE}{clip(round(0.4742117275,4))}
\FPeval{\StarRatingRecallAE}{clip(round(0.6895098766,4))}

\FPeval{\PositivePrecisionAE}{clip(round(0.8571785355,4))}
\FPeval{\NegativePrecisionAE}{clip(round(0.8053341237,4))}
\FPeval{\StarRatingPrecisionAE}{clip(round(0.2297734887,4))}

\FPeval{\PositiveFAE}{clip(round(0.7064112664,4))}
\FPeval{\NegativeFAE}{clip(round(0.4165524884,4))}
\FPeval{\StarRatingFAE}{clip(round(0.2994241075,4))}

\begin{table*}[htb]
\renewcommand{\arraystretch}{1.5}
\centering
\caption{Performance Measures of Star Rating Vector Predictions}
\label{Table:AmznVecPredError}

\begin{tabularx}{\textwidth}{
|>{\centering\arraybackslash}X
|>{\centering\arraybackslash}X
|>{\centering\arraybackslash}X
|>{\centering\arraybackslash}X
|>{\centering\arraybackslash}X
|>{\centering\arraybackslash}X
|}
\hline
 \multirow{2}{*}{\textbf{Train set (\%)}} &
 \multirow{2}{*}{\textbf{Category}} &
 \multicolumn{4}{c|}{\textbf{Performance Measure}} \\
 \hhline{~~----}
 & & \textbf{Accuracy} & \textbf{Recall} & \textbf{Precision} & \textbf{F1 Score}\\
 \hline
 \multirow{3}{*}{95} & Positive & 0.5406 & 0.7496 & 0.8601 & 0.7068\\
 & Negative & 0.2062 & 0.4775 & 0.8067 & 0.4207\\
 & Star Rating & 0.6930 & 0.6912 & 0.2259 & 0.2921\\
 \hline
 \multirow{3}{*}{90} & Positive & 0.5420 & 0.7524 & 0.8589 & 0.7088\\
 & Negative & 0.2052 & 0.4753 & 0.8069 & 0.4192\\
 & Star Rating & 0.6931 & 0.6913 & 0.2267 & 0.2945\\
 \hline
 \multirow{3}{*}{80} & Positive & 0.5416 & 0.7527 & 0.8571 & 0.7075\\
 & Negative & 0.2038 & 0.4718 & 0.8077 & 0.4159\\
 & Star Rating & 0.6917 & 0.6896 & 0.2236 & 0.2912\\
 \hline
 \multirow{3}{*}{70} & Positive & 0.5410 & 0.7503 & 0.8588 & 0.7065\\
 & Negative & 0.2046 & 0.4751 & 0.8051 & 0.4176\\
 & Star Rating & 0.6925 & 0.6905 & 0.2280 & 0.2975\\
 \hline
 \multirow{3}{*}{50} & Positive & 0.5403 & 0.7514 & 0.8572 & 0.7064\\
 & Negative & 0.2040 & 0.4742 & 0.8053 & 0.4166\\
 & Star Rating & 0.6915 & 0.6895 & 0.2298 & 0.2994\\
 \hline
\end{tabularx}
\end{table*}

\section{Results}
\label{Sec:Res}

Table~\ref{Table:VecPredMeasure} shows the results obtained for the preference measure defined in Section~\ref{sec:eval} for the \textit{Core Reaction Set Model} introduced in Section~\ref{sec:core} and \textit{All Reaction Set Model} introduced in Section~\ref{sec:all}. 
All reactions except \textit{Sad} reach their highest F1 score at the $95\%-05\%$ train-test division, while the \textit{Sad} reaction reaches its peak F1 score at the $80\%-20\%$ division.
Interestingly, the performance of the model in predicting each reaction seems to roughly follow a specific pattern; reactions that were used more often in the data set seem to have a higher F1 score than reactions that were used less often, with the exception of the F1 score of \textit{Wow} being higher than that of \textit{Sad}.
Figure~\ref{fig:Err} portrays the F1 score for each reaction as the train-test division varies for the \textit{Core Reaction Set Model}.
In the case of \textit{All Reaction Set Model}, as shown in Table~\ref{Table:VecPredMeasure}, while the F1 of \textit{Like} was much higher than that of other reactions, its inclusion brought forth significant reductions in the F1 scores of the other reactions. The \textit{Thankful} reaction had a F1 of almost zero.

\pgfplotsset{compat=1.15}

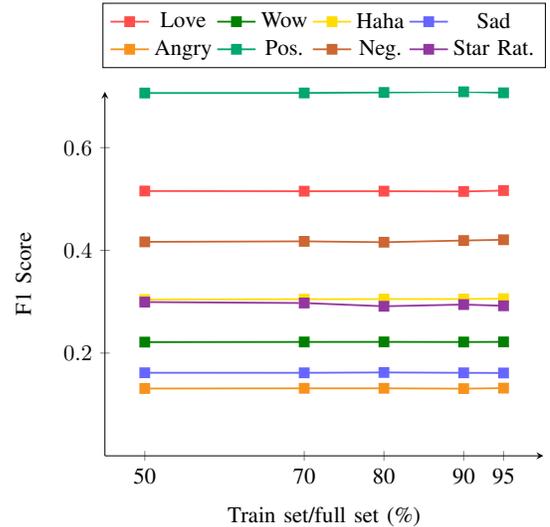
\begin{figure}[!htb]
\centering
\resizebox{0.4\textwidth}{!}{%
\begin{tikzpicture}
\pgfplotsset{
every axis legend/.append style={
at={(0.5,1.03)},
anchor=south
},
}
\begin{axis}[
    legend columns=4,
    axis lines=middle,
    xmin=45, 
    xmax=100,
    ymin=0,
    x label style={
        at={(axis description cs:0.5,-0.03)},
        below=5mm},
    y label style={
        at={(axis description cs:-0.15,0.5)},
        rotate=90,
        anchor=south},
    legend style={above=2mm},
    xlabel=Train set/full set (\%) ,
    ylabel=F1 Score,
    enlargelimits = false,
    xticklabels from table={Err.dat}{Division},xtick=data]
\addplot[red!70,thick,mark=square*] table [y=Love,x=X]{Err.dat};
\addlegendentry{Love}
\addplot[green!50!black,thick,mark=square*] table [y=Wow,x=X]{Err.dat};
\addlegendentry{Wow}
\addplot[yellow!90!red,thick,mark=square*] table [y=Haha,x=X]{Err.dat};
\addlegendentry{Haha}
\addplot[blue!60,thick,mark=square*] table [y=Sad,x=X]{Err.dat};
\addlegendentry{Sad}
\addplot[yellow!50!red,thick,mark=square*] table [y=Angry,x=X]{Err.dat};
\addlegendentry{Angry}
\addplot[cyan!20!green,thick,mark=square*] table [y=Positive,x=X]{Err.dat};
\addlegendentry{Pos.}
\addplot[red!20!brown,thick,mark=square*] table [y=Negetive,x=X]{Err.dat};
\addlegendentry{Neg.}
\addplot[white!30!purple!70!blue,thick,mark=square*] table [y=StarRating,x=X]{Err.dat};
\addlegendentry{Star Rat.}
\end{axis}
\end{tikzpicture}
}
\caption{Change of the F1 score of the Core Reaction Model and the Star Rating Model with Train-Test Division}
\label{fig:Err}
\end{figure}

The results obtained for \textit{Star Rating Model} introduced in section~\ref{sec:star} are shown in table~\ref{Table:AmznVecPredError}. In contrast to the results obtained for \textit{Positive} and \textit{Negative} components, aggregation of reactions into a single Star Rating value has caused a significant decrease in precision; possibly due to the discrete nature of the Star Rating value which is divided into bins at 0.5 intervals. Figure~\ref{fig:Err} portrays the change of F1 value with the train-test division.

As portrayed by Figure~\ref{fig:Err}, the performance of the models remains largely unaffected by the train-test division chosen. The reason could be the large size of the data set; the number of unique words in the train data set does not change significantly for different train-test divisions.

\section{Conclusion}
\label{Sec:Con}
Upon comparing the \textit{Star Rating Model} with the \textit{Core Reaction Set Model}, it becomes evident that the F1 scores are significantly improved upon the accumulation of separate reaction values into two categories as \textit{Positive} and \textit{Negative}. A possible reason for this is the possibility of the intra-category measurement errors being eliminated due to merging. However, merging all reactions into a single Star Rating value accentuates errors. This could be accounted to the additional error margin introduced by discretization. The negative effect of \textit{Like} and \textit{Thankful} reactions, which were eliminated in the \textit{Core Reaction Set Model} due to their abnormal counts, could be proven as well. The inclusion of those reactions caused significant reductions in the F1 scores of the other reactions as can be seen from the results of the \textit{All Reaction Set Model}.

This study represents modelling efforts that may be considered classical and limited in nature. \citet{kowsari2019text} highlights a number of pre-processing steps and algorithms that may be combined with the feature engineering work presented here for potentially more accurate models in the future. As noted therein, deep learning techniques hold particular promise. Due to limited or missing language resources and tooling, as noted by \citet{de2019survey}, some pre-processing techniques may not be possible in Sinhala. Building these tools may further increase the accuracy even with a simplistic model.

\bibliographystyle{IEEEtranN}
\bibliography{bibliography.bib}

\end{document}